\documentclass{article}

     \PassOptionsToPackage{numbers, compress}{natbib}

     \usepackage[final]{neurips_2024}

\usepackage[utf8]{inputenc} 
\usepackage[T1]{fontenc}    
\usepackage{hyperref}       
\usepackage{url}            
\usepackage{booktabs}       
\usepackage{amsfonts}       
\usepackage{nicefrac}       
\usepackage{microtype}      
\usepackage{xcolor}         
\usepackage{amsmath}        
\usepackage{graphicx}
\usepackage{multirow}
\usepackage{makecell}

\title{A Study on Educational Data Analysis and Personalized Feedback Report Generation Based on Tags and ChatGPT\thanks{Zhou, Y., Zhang, M., Jiang, Y.-H., Gao, X., Liu, N., \& Jiang, B. (2024). A Study on Educational Data Analysis and Personalized Feedback Report Generation Based on Tags and ChatGPT. Conference Proceedings of the 28th Global Chinese Conference on Computers in Education (GCCCE 2024), 108–115. Chongqing, China: Global Chinese Conference on Computers in Education.}}
\author{
    Yizhou Zhou$^{1, 2, 3, 4}$\thanks{Corresponding Author: zhouyizhou25@stu.ecnu.edu.cn}, 
    Mengqiao Zhang$^{5}$, 
    Yuan-Hao Jiang$^{2, 3, 4}$, 
    Xinyu Gao$^{6}$, 
    Naijie Liu$^{7}$, 
    Bo Jiang$^{2,3, 4}$ \\
    \\
    $^{1}$~School of Design and Engineering, National University of Singapore \\
    $^{2}$~Lab for Artificial Intelligence for Education, East China Normal University \\
    $^{3}$~Shanghai Institute of Artificial Intelligence for Education, East China Normal University \\
    $^{4}$~School of Computer Science and Technology, East China Normal University \\
    $^{5}$~National Institute of Education, Nanyang Technological University \\
    $^{6}$~College of Education Science and Technology, Zhejiang University of Technology \\
    $^{7}$~School of Management Science and Engineering, Shandong University of Finance and Economics \\
}

\begin{document}
\maketitle
\begin{abstract}
This study introduces a novel method that employs tag annotation coupled with the ChatGPT language model to analyze student learning behaviors and generate personalized feedback. Central to this approach is the conversion of complex student data into an extensive set of tags, which are then decoded through tailored prompts to deliver constructive feedback that encourages rather than discourages students. This methodology focuses on accurately feeding student data into large language models and crafting prompts that enhance the constructive nature of feedback. The effectiveness of this approach was validated through surveys conducted with over 20 mathematics teachers, who confirmed the reliability of the generated reports. This method can be seamlessly integrated into intelligent adaptive learning systems or provided as a tool to significantly reduce the workload of teachers, providing accurate and timely feedback to students. By transforming raw educational data into interpretable tags, this method supports the provision of efficient and timely personalized learning feedback that offers constructive suggestions tailored to individual learner needs.
\end{abstract}

\section{Introduction}

The development of Large Language Models (LLMs) such as GPT-4 has significantly impacted various domains, especially education \cite{doughty_comparative_2024}. These models are adept at understanding and generating natural language, offering significant capabilities in text generation, question-answering, and more \cite{liu_chatqa_2024}. In particular in education, LLMs are poised to enhance learning experiences and provide immediate and constructive feedback to students. However, effectively integrating the extensive and complex data typical of educational settings into these models presents notable challenges \cite{keuning_systematic_2019}.

Immediate feedback is a strong reinforcement mechanism, underscored by Skinner's operant conditioning theory, which asserts that immediate rewards or corrections can significantly influence future behavior by reinforcing desired learning activities \cite{skinner_operant_1963,arifin_application_2021}. This is particularly vital in educational contexts, where timely feedback can help students quickly recognize and correct mistakes, fostering a better understanding and retention of material \cite{hao_towards_2022}. In addition, personalized feedback according to individual learning trajectories allows personalized educational guidance, making learning more effective.

However, while immediate feedback is invaluable, the deployment of LLMs such as GPT-4 to generate comprehensive educational feedback presents significant challenges, particularly in data handling. Traditional LLM applications tend to focus on singular tasks, such as grading individual questions or essays, providing feedback limited to specific prompts without a broader contextual understanding. This narrow approach often misses the need for cumulative feedback that offers constructive insight over a learning period. To address this, we propose a tag annotation method that refines raw educational data into a structured format of predefined tags. These tags represent essential attributes of the student data comprehensively, enabling the model to process vast information more efficiently and generate feedback that is both integrative and strategically supportive of the student's educational journey. This method allows for the production of rich, context-aware feedback that supports holistic student development, overcoming the limitations of traditional single-point assessments.

Our research utilizes data from an adaptive learning system tested in a Shanghai primary school, captures detailed records of students’ performances across various parameters including correctness, difficulty levels, knowledge categories, ability levels, and task completion time. By transforming these data points into tags, we enable the LLM to generate nuanced feedback. This feedback is not only precise but also supportive, encouraging students to engage positively with their learning material. This approach of tag annotation simplifies the model's data processing tasks and enhances the relevance of its outputs, making LLMs more practical for real-world educational applications. It marks a significant advance in the integration of advanced AI technologies into education, paving the way for scalable and impactful personalized learning feedback.

\section{Related Work}

\subsection{Traditional Feedback Generation with LLMs in Education}

The integration of Large Language Models like ChatGPT into education has highlighted the potential for automated feedback. Traditionally, such models provide immediate responses to individual tasks, such as answers to questions, essays, or coding projects \cite{hao_towards_2022}. However, they have yet to fully address the challenge of analyzing large datasets for more comprehensive feedback. There is a growing need for methods that can efficiently encode broader educational data into these models. This would enable the generation of detailed, personalized feedback reports reflecting the student's learning journey over time.

\subsection{Effective Reporting in Educational Feedback}

Research indicates that the most effective feedback reports are those that provide holistic insights into a student's learning progress, addressing both strengths and areas for improvement. Kobus (2007) found that reports combining detailed analysis with positive reinforcement encourage students more effectively \cite{kobus_increasing_2007}. These studies underscore the importance of balanced feedback that motivates students while guiding them toward academic improvement. Periodic, comprehensive feedback reports are more beneficial as they allow for adjustments in teaching strategies and student learning approaches, aligning more closely with ongoing educational needs \cite{lyon_capabilities_2016}. Additionally, the tone of the feedback is crucial; emphasizing constructive and supportive language helps maintain a positive learning environment \cite{kobus_increasing_2007}. Avoiding overly critical or negative remarks is essential to prevent discouraging students. Nurturing a student's confidence and interest in learning through supportive feedback significantly enhances learning outcomes \cite{henderson_developing_2005}, which not only corrects misconceptions but also bolsters the student's abilities and self-esteem, fostering a conducive atmosphere for continuous learning and growth.

\section{Experiment and Data Analysis}

\subsection{Experimental Procedure and Data Acquisition}

The study collected experimental data from two primary school classes in Shanghai, utilizing a three-dimensional adaptive learning system developed by the Lab for Artificial Intelligence in Education at East China Normal University.This system addresses knowledge, ability, and affective attitude, providing a variety of multiple-choice questions, each with two possible outcomes: correct (1) and incorrect (0). Students engaged in the system's online learning sessions during designated class periods on Monday and Wednesday afternoons each week.
\begin{figure}
    \centering
    \includegraphics[width=0.5\linewidth]{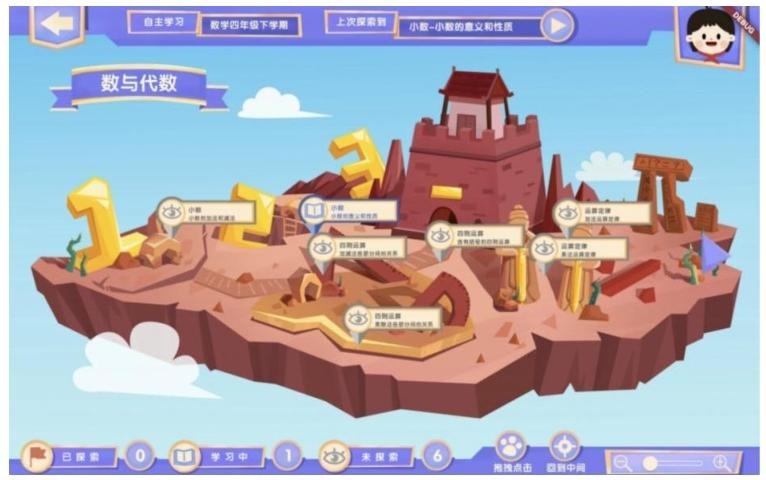}
    \caption{Adaptive learning system interface display.}
    \label{fig:enter-label}
\end{figure}
The experiment was carried out within a real online educational platform, specifically the aforementioned adaptive learning system, which included advanced data tracking modules. This setup not only ensured the accurate collection of comprehensive learning data but also strictly adhered to data protection measures to safeguard student privacy. Specifically, the final dataset provided by the system allowed for the careful extraction of data on students' performance in multiple-choice questions, including their accuracy rates. Additionally, we gathered data on the knowledge categories, ability levels associated with each question, and the time students took to complete these questions. Conducting the experiment during afternoon class periods ensured students were engaged and not distracted by other academic responsibilities, thus reflecting their true learning capabilities more accurately. This careful arrangement provided a solid foundation for subsequent analyses, aiming to evaluate the effectiveness of tag-based feedback and adaptive learning strategies employed.

\subsection{Data Processing and Tag Annotation}

Effective use of educational data in generating meaningful feedback through Large Language Models (LLMs) like GPT-4 requires thorough preprocessing and organization of raw data into a structured and interpretable format. The initial dataset included a vast array of knowledge categories and ability levels—more than fifty and thirty distinct types, respectively. To enhance model practicality, these were consolidated into manageable classes.

Knowledge categories were refined into six primary groups: Calculations of Speed and Time, Geometric Shapes and Properties, Data Statistics and Probability, Algebra and Functions, Arithmetic Operations and Properties. This categorization aimed to cover a broad spectrum of subjects while simplifying complexity. Ability levels were similarly reduced and organized into six groups: Practical Mathematical Application Skills, Data Organization and Statistical Skills, Computational Skills, Geometric Thinking Skills, Reasoning and Logical Thinking, and Innovative and Abstract Thinking. These groups reflected the core skills essential for academic success in structured learning environments.

A major challenge was managing multiple entries per student per question, often including brief, non-essential attempts likely from navigational actions rather than genuine problem-solving efforts. To resolve this, only the longest duration of attempt per question was retained, considered to represent the most substantial effort. Additionally, entries with zero duration were discarded, presumed to be guesses or inactive interactions, thus providing little educational value.

The streamlined data led to the development of a tag annotation system, organized into three primary categories: Performance Tags, Knowledge Domain Tags, and Ability Level Tags. Performance Tags (12 tags) were based on a combination of difficulty level (1-3), accuracy, and speed. Accuracy was split into 'adequate' (above 65\%) and 'struggling' (below 55\%). Speed was assessed by comparing each student's performance against their peers; top or bottom performers were tagged if the sample size exceeded 40, otherwise, the fastest or slowest 50\% received the tag. Knowledge Domain Tags (10 tags) reflected proficiency or challenges in each of the five knowledge areas, with each capable of producing a positive or negative tag based on performance relative to benchmarks. Ability Level Tags (12 tags) assessed whether students excelled or struggled in each of the six domains.Specific tag settings are detailed in Table 1.

\begin{table}[ht]
\centering
\caption{Labels correspond to data features.}
\label{tab:data_labels}
\begin{tabular}{|p{2.2cm}|p{1.5cm}|p{9cm}|}
\hline
\textbf{Tag Category}       & \textbf{Tag}       & \textbf{Description}                                                                                   \\ \hline
\multirow{12}{*}{\makecell[l]{Basic\\Analysis}} 
                            & Tag\_1\_1          & Correctly and quickly on easy questions.                                                              \\  
                            & Tag\_1\_2          & Correctly and quickly on medium difficulty questions.                                                 \\  
                            & Tag\_1\_3          & Correctly and quickly on difficult questions.                                                         \\  
                            & Tag\_1\_4          & Correctly but completed slowly on easy questions.                                                     \\  
                            & Tag\_1\_5          & Correctly but completed slowly on medium difficulty questions.                                        \\  
                            & Tag\_1\_6          & Correctly but completed slowly on difficult questions.                                                \\  
                            & Tag\_1\_7          & Incorrectly but completed quickly on easy questions.                                                  \\  
                            & Tag\_1\_8          & Answered incorrectly but quickly on medium difficulty questions.                                      \\  
                            & Tag\_1\_9          & Incorrectly but completed quickly on difficult questions.                                             \\  
                            & Tag\_1\_10         & Incorrectly but completed slowly on easy questions.                                                   \\  
                            & Tag\_1\_11         & Incorrectly but completed slowly on medium difficulty questions.                                      \\  
                            & Tag\_1\_12         & Incorrectly but completed slowly on difficult questions.                                              \\ \hline
\multirow{10}{*}{\makecell[l]{Knowledge\\Category\\Analysis}} 
                            & Tag\_2\_1          & Outstanding performance in calculations speed and time.                                     \\  
                            & Tag\_2\_2          & Outstanding in the identification and geometric shapes.                                 \\  
                            & Tag\_2\_3          & Outstanding in data statistics and probability problems.                                              \\  
                            & Tag\_2\_4          & Outstanding in algebraic equations and functions.                                                     \\  
                            & Tag\_2\_5          & Outstanding in arithmetic operations and properties.                                                  \\  
                            & Tag\_2\_6          & Struggling with calculations involving speed and time.                                                \\  
                            & Tag\_2\_7          & Struggling to recognize and work with geometric shapes.                                 \\  
                            & Tag\_2\_8          & Finding data statistics and probability problems challenging.                                         \\  
                            & Tag\_2\_9          & Struggling with algebraic equations and functions.                                                    \\  
                            & Tag\_2\_10         & Struggling with arithmetic operations and properties.                                                 \\ \hline
\multirow{12}{*}{\makecell[l]{Ability\\Analysis}} 
                            & Tag\_3\_1          & Strong capabilities in practical application of mathematics.                                          \\  
                            & Tag\_3\_2          & Strong capabilities in statistical analysis.                                                          \\  
                            & Tag\_3\_3          & Strong computational skills.                                                                          \\  
                            & Tag\_3\_4          & Strong geometric thinking skills.                                                                     \\  
                            & Tag\_3\_5          & Strong logical reasoning skills.                                                                      \\  
                            & Tag\_3\_6          & Strong innovative and abstract thinking skills.                                                       \\  
                            & Tag\_3\_7          & Challenged by the practical application of mathematics.                                               \\  
                            & Tag\_3\_8          & Challenged by statistical analysis.                                                                   \\  
                            & Tag\_3\_9          & Challenged by computational skills.                                                                   \\  
                            & Tag\_3\_10         & Challenged by geometric thinking skills.                                                              \\  
                            & Tag\_3\_11         & Challenged by logical reasoning.                                                                      \\  
                            & Tag\_3\_12         & Challenged by innovative and abstract thinking.                                                       \\ \hline
\end{tabular}
\end{table}

In total, 34 tags were designed to cover the potential spectrum of student performance scenarios comprehensively. This system ensured that each student could be associated with multiple tags, providing a detailed picture of their learning behaviors and outcomes, critical for generating personalized educational feedback.

\subsection{Tag Parsing and Report Generation}

The conclusion of our methodology involves transforming processed data into structured tags and combining these with the GPT-4 model to generate personalized educational feedback reports. After data refinement and tag annotation as outlined in the previous sections, we create a complete dataset named \texttt{student\_tag}. This dataset stores each student's 34 identified tags, where each tag is denoted as either 0 or 1, indicating the absence or presence of a specific learning characteristic or challenge.

The process of Prompt Design begins with the retrieval of student tags using the \texttt{get\_student\_tags} function from the \texttt{student\_tag} dataset. These tags provide customized inputs for generating reports, ensuring that each student's feedback is uniquely tailored to their individual performance data. The prompts designed for interaction with GPT-4 are carefully structured to include several distinct sections: an Overview, Basic Analysis, Knowledge Category Analysis, Ability Analysis, Learning Strategies and Recommendations, and a Summary. This structured approach aids the model in understanding the purpose of each section and in generating corresponding content. Each segment of the prompt includes specific instructions and guidelines that outline the type of content needed and its intended purpose, helping the model produce more accurate and relevant text outputs. The use of "you" in the prompts to directly address the student enhances the relatability and specificity of the reports, fostering greater engagement from the student.

The use of GPT-4 employs \texttt{openai.ChatCompletion.create} to utilize the model. Control over the randomness and length of the generated text is achieved by setting the \texttt{Temperature} to 0.4, which ensures that the generated text is consistent and minimally random, suitable for the needs of a professional report \cite{zong_solving_2023}. The \texttt{Max\_tokens} is set to 1000 to allow ample space for a thorough report. The \texttt{Top\_p} is configured to 1, utilizing the full potential vocabulary to enrich the text's comprehensiveness and coverage. Controls on repetition and innovation, such as \texttt{frequency\_penalty} and \texttt{presence\_penalty}, are set to 0. This decision indicates a preference to maintain the natural flow and consistency of the content, which is deemed more crucial in this context.

For practical illustration, a sample report for randomly selected student number 2965 is provided in the appendix. This example demonstrates the tailored feedback generated by the system.

Through these mechanisms, the system effectively synthesizes complex educational data into actionable insights, providing students with constructive feedback that is both encouraging and tailored to promote educational advancement and personal growth.

\section{Evaluation}

The potential of large language models (LLMs) like GPT-4 for educational feedback has been demonstrated through an evaluation with primary level mathematics teachers. The assessment, conducted via a questionnaire, examined five key aspects of the generated reports - comprehensibility, practicality, motivation, clarity, and organizational structure. Each aspect was rated on a scale from 0 to 10, with a maximum score of 50.

We received 63 questionnaires, discarding three for unjustifiably low scores and another 28 for non-genuine perfect scores, leaving 32 valid responses for analysis. Overall, the average scores exceeded 7 points in most dimensions, indicating that the reports were well-received. Teachers found them helpful for understanding and supporting the learning process, particularly in identifying student strengths and challenges.

The evaluation highlighted the need for simpler language and better data presentation, as clarity received the lowest average score of 6.59. A boxplot analysis revealed varying teacher perceptions across different aspects. Understanding Level had a median score of 7.5, indicating effective capture of students' strengths and challenges. Practicality scored slightly lower, reflecting varied teacher opinions on the suggestions' applicability. The Motivation effect, with a median near 8, showed a positive impact, despite a few low outliers. Clarity and Organizational Structure had median scores of around 7 and above 7.5 respectively, suggesting the need for clearer reports and confirming the reports' logical structure facilitated information assimilation.

\begin{table}[ht]
\centering
\caption{Questionnaire items for the assessment of personalized feedback reports.}
\label{tab:questionnaire}
\begin{tabular}{|p{3cm}|p{9.5cm}|}
\hline
\textbf{Category} & \textbf{Question Description} \\ \hline
Understanding Level & Do you believe this report accurately analyzes students' main strengths and challenges in their learning? \\ \hline
Practicality & Are the recommendations proposed in the report practical and feasible? Do you think these suggestions are easily applicable in actual teaching practices? \\ \hline
Motivation effect & Do you feel that this report effectively motivates students to improve their learning methods and enhance their learning attitudes? \\ \hline
Clarity & Is the language used in the report clear? Did you encounter any parts that were difficult to understand while reading it? \\ \hline
Organizational structure & Do you find the structure of the report to be logical and conducive to quickly grasping key information? Are the various sections of the report well-organized and easy to follow? \\ \hline
Advise & Do you have any additional suggestions for this report? \\ \hline
\end{tabular}
\end{table}

In summary, teachers affirmed the reports' effectiveness in highlighting understanding, practicality, and organizational structure, with a recommendation to improve clarity to avoid misinterpretations. Such insights are crucial as they guide the refinement of algorithms that underpin report generation. This reaffirms the viability of using LLMs like GPT-4 to provide personalized educational feedback, enhancing the learning experience.

\section{Discussion and Implications}

This study has demonstrated the potential of the GPT-4 large language model in generating personalized educational feedback reports in the field of education. Most teachers gave positive reviews of the learning reports produced, particularly in terms of analyzing students' learning strengths, challenges, and providing specific suggestions. These feedback reports can serve as a powerful tool to help teachers understand and support students' learning processes more effectively, echoing the importance of immediate feedback in education.

\begin{figure}
    \centering
    \includegraphics[width=1\linewidth]{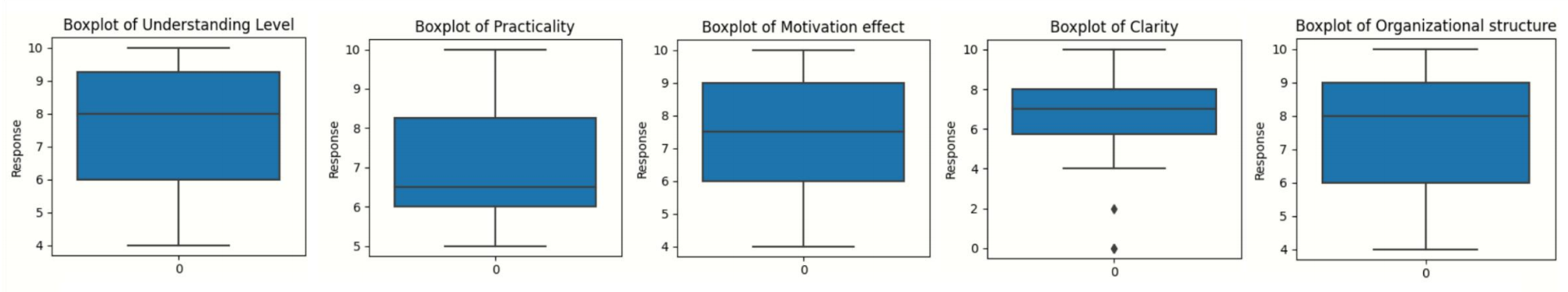}
    \caption{Boxplot of questionnaire results across five dimensions.}
    \label{fig:enter-label}
\end{figure}

The lower evaluations for report clarity highlight areas for improvement, suggesting that future report generation should focus more on the simplicity of language and the clarity of data presentation to ensure that both teachers and students can understand them better. Additionally, the presence of low outliers also indicates potential shortcomings in how the reports motivate students to improve their learning methods.

The significance of this study lies in providing a new perspective and method for using AI technology to support personalized educational feedback. Through the process of tag annotation and tag parsing, it is possible to transform vast amounts of complex learning behavior data into comprehensive and accurate input.

Future adjustments to the prompts can also be made to ensure the report outputs better meet the expectations and needs of math teachers in the classroom, effectively helping to reduce teacher workload and providing students with timely, personalized analysis of their learning status.

\section{Conclusion}
This study explored a method of using tag annotation and tag parsing to implement the large language model ChatGPT in generating personalized, real-time educational feedback reports, confirming its potential to play a key role in personalized education. By converting students' multidimensional learning data into structured tags, the model is capable of generating precise, personalized learning reports. These reports not only summarize students' performance but also provide specific strategies for improvement.

Feedback from teachers indicates that such reports are substantially helpful in understanding students' learning situations and providing targeted guidance. However, variations in scores for clarity and motivational impact also reveal areas for improvement, pointing out directions for future research and development.

In summary, this research provides a successful case study demonstrating the practicality and effectiveness of LLMs in the field of education, offering valuable insights and guidance for future AI research in similar areas. It not only reinforces the concept of data-driven educational feedback but also opens new avenues for educators and researchers to further explore the role of artificial intelligence in enhancing teaching quality and the learning experience.

\section*{Acknowledgements}

This work was partially supported by the National Natural Science Foundation of China under Grant 61977058, and the Natural Science Foundation of Shanghai under Grant 23ZR1418500.

\bibliographystyle{IEEEtran}
\bibliography{GCCCE2024论文参考文献}

\begin{thebibliography}{10}
\providecommand{\url}[1]{#1}
\csname url@samestyle\endcsname
\providecommand{\newblock}{\relax}
\providecommand{\bibinfo}[2]{#2}
\providecommand{\BIBentrySTDinterwordspacing}{\spaceskip=0pt\relax}
\providecommand{\BIBentryALTinterwordstretchfactor}{4}
\providecommand{\BIBentryALTinterwordspacing}{\spaceskip=\fontdimen2\font plus
\BIBentryALTinterwordstretchfactor\fontdimen3\font minus \fontdimen4\font\relax}
\providecommand{\BIBforeignlanguage}[2]{{%
\expandafter\ifx\csname l@#1\endcsname\relax
\typeout{** WARNING: IEEEtran.bst: No hyphenation pattern has been}%
\typeout{** loaded for the language `#1'. Using the pattern for}%
\typeout{** the default language instead.}%
\else
\language=\csname l@#1\endcsname
\fi
#2}}
\providecommand{\BIBdecl}{\relax}
\BIBdecl

\bibitem{doughty_comparative_2024}
\BIBentryALTinterwordspacing
J.~Doughty, Z.~Wan, A.~Bompelli, J.~Qayum, T.~Wang, J.~Zhang, Y.~Zheng, A.~Doyle, P.~Sridhar, A.~Agarwal, C.~Bogart, E.~Keylor, C.~Kultur, J.~Savelka, and M.~Sakr, ``\BIBforeignlanguage{en}{A {Comparative} {Study} of {AI}-{Generated} ({GPT}-4) and {Human}-crafted {MCQs} in {Programming} {Education}},'' in \emph{\BIBforeignlanguage{en}{Proceedings of the 26th {Australasian} {Computing} {Education} {Conference}}}.\hskip 1em plus 0.5em minus 0.4em\relax Sydney NSW Australia: ACM, Jan. 2024, pp. 114--123. [Online]. Available: \url{https://dl.acm.org/doi/10.1145/3636243.3636256}
\BIBentrySTDinterwordspacing

\bibitem{liu_chatqa_2024}
\BIBentryALTinterwordspacing
Z.~Liu, W.~Ping, R.~Roy, P.~Xu, M.~Shoeybi, and B.~Catanzaro, ``Chatqa: {Building} gpt-4 level conversational qa models,'' \emph{arXiv e-prints}, pp. arXiv--2401, 2024. [Online]. Available: \url{https://ui.adsabs.harvard.edu/abs/2024arXiv240110225L/abstract}
\BIBentrySTDinterwordspacing

\bibitem{keuning_systematic_2019}
\BIBentryALTinterwordspacing
H.~Keuning, J.~Jeuring, and B.~Heeren, ``\BIBforeignlanguage{en}{A {Systematic} {Literature} {Review} of {Automated} {Feedback} {Generation} for {Programming} {Exercises}},'' \emph{\BIBforeignlanguage{en}{ACM Transactions on Computing Education}}, vol.~19, no.~1, pp. 1--43, Mar. 2019. [Online]. Available: \url{https://dl.acm.org/doi/10.1145/3231711}
\BIBentrySTDinterwordspacing

\bibitem{skinner_operant_1963}
\BIBentryALTinterwordspacing
B.~F. Skinner, ``Operant behavior.'' \emph{American psychologist}, vol.~18, no.~8, p. 503, 1963, publisher: American Psychological Association. [Online]. Available: \url{https://psycnet.apa.org/journals/amp/18/8/503/}
\BIBentrySTDinterwordspacing

\bibitem{arifin_application_2021}
\BIBentryALTinterwordspacing
Z.~Arifin and H.~Humaedah, ``Application of {Theory} {Operant} {Conditioning} {BF} {Skinner}'s in {PAI} {Learning}: {Penerapan} {Teori} {Operant} {Conditioning} {BF} {Skinner} {Dalam} {Pembelajaran} {PAI},'' \emph{Journal of Contemporary Islamic Education}, vol.~1, no.~2, pp. 101--110, 2021. [Online]. Available: \url{https://journal.iaimnumetrolampung.ac.id/index.php/cie/article/view/1602}
\BIBentrySTDinterwordspacing

\bibitem{hao_towards_2022}
\BIBentryALTinterwordspacing
Q.~Hao, D.~H. Smith~Iv, L.~Ding, A.~Ko, C.~Ottaway, J.~Wilson, K.~H. Arakawa, A.~Turcan, T.~Poehlman, and T.~Greer, ``\BIBforeignlanguage{en}{Towards understanding the effective design of automated formative feedback for programming assignments},'' \emph{\BIBforeignlanguage{en}{Computer Science Education}}, vol.~32, no.~1, pp. 105--127, Jan. 2022. [Online]. Available: \url{https://www.tandfonline.com/doi/full/10.1080/08993408.2020.1860408}
\BIBentrySTDinterwordspacing

\bibitem{kobus_increasing_2007}
\BIBentryALTinterwordspacing
T.~Kobus, L.~Maxwell, and J.~Provo, ``Increasing {Motivation} of {Elementary} and {Middle} {School} {Students} through {Positive} {Reinforcement}, {Student} {Self}-{Assessment}, and {Creative} {Engagement}.'' \emph{Online Submission}, 2007, publisher: ERIC. [Online]. Available: \url{https://eric.ed.gov/?id=ED498971}
\BIBentrySTDinterwordspacing

\bibitem{lyon_capabilities_2016}
\BIBentryALTinterwordspacing
A.~R. Lyon, C.~C. Lewis, M.~R. Boyd, E.~Hendrix, and F.~Liu, ``\BIBforeignlanguage{en}{Capabilities and {Characteristics} of {Digital} {Measurement} {Feedback} {Systems}: {Results} from a {Comprehensive} {Review}},'' \emph{\BIBforeignlanguage{en}{Administration and Policy in Mental Health and Mental Health Services Research}}, vol.~43, no.~3, pp. 441--466, May 2016. [Online]. Available: \url{http://link.springer.com/10.1007/s10488-016-0719-4}
\BIBentrySTDinterwordspacing

\bibitem{henderson_developing_2005}
\BIBentryALTinterwordspacing
P.~Henderson, A.~C. Ferguson-Smith, and M.~H. Johnson, ``\BIBforeignlanguage{en}{Developing essential professional skills: a framework for teaching and learning about feedback},'' \emph{\BIBforeignlanguage{en}{BMC Medical Education}}, vol.~5, no.~1, p.~11, Apr. 2005. [Online]. Available: \url{https://bmcmededuc.biomedcentral.com/articles/10.1186/1472-6920-5-11}
\BIBentrySTDinterwordspacing

\bibitem{zong_solving_2023}
\BIBentryALTinterwordspacing
M.~Zong and B.~Krishnamachari, ``Solving math word problems concerning systems of equations with gpt-3,'' in \emph{Proceedings of the {AAAI} {Conference} on {Artificial} {Intelligence}}, vol.~37, 2023, pp. 15\,972--15\,979, issue: 13. [Online]. Available: \url{https://ojs.aaai.org/index.php/AAAI/article/view/26896}
\BIBentrySTDinterwordspacing

\end{thebibliography}

\end{document}